\pdfoutput=1
\documentclass[10pt,twocolumn,letterpaper]{article}

\usepackage[pagenumbers]{cvpr} 

\usepackage{graphicx}
\usepackage{amsmath}
\usepackage{amssymb}
\usepackage{booktabs}
\usepackage{graphicx}
\usepackage{multirow}
\usepackage{makecell}
\usepackage{arydshln}
\usepackage[export]{adjustbox}
\usepackage[accsupp]{axessibility}  

\usepackage{pifont}
\usepackage{xcolor}
\definecolor{deemph}{gray}{0.6}

\usepackage[pagebackref,breaklinks,colorlinks]{hyperref}

\usepackage[capitalize]{cleveref}
\crefname{section}{Sec.}{Secs.}
\Crefname{section}{Section}{Sections}
\Crefname{table}{Table}{Tables}
\crefname{table}{Tab.}{Tabs.}

\def\cvprPaperID{130} 
\def\confName{CVPR}
\def\confYear{2022}

\begin{document}

\title{DMTNet: Dynamic Multi-scale Network for Dual-pixel Images Defocus Deblurring with Transformer}

\author{ Dafeng Zhang$^{1}$\thanks{Equal contribution.}\hspace{20pt}  Xiaobing Wang$^1$\footnotemark[1]\\ 
{$^1 $} Samsung Research China - Beijing (SRC-B) \\
{\tt\small \{dafeng.zhang, x0106.wang\}@samsung.com}
}
\maketitle

\maketitle

\begin{abstract}
	Recent works achieve excellent results in defocus deblurring task based on dual-pixel data using convolutional neural network (CNN), while the scarcity of data limits the exploration and attempt of vision transformer in this task. In addition, the existing works use fixed parameters and network architecture to deblur images with different distribution and content information, which also affects the generalization ability of the model. In this paper, we propose a dynamic multi-scale network, named DMTNet, for dual-pixel images defocus deblurring. DMTNet mainly contains two modules: feature extraction module and reconstruction module. The feature extraction module is composed of several vision transformer blocks, which uses its powerful feature extraction capability to obtain richer features and improve the robustness of the model. The reconstruction module is composed of several Dynamic Multi-scale Sub-reconstruction Module (DMSSRM). DMSSRM can restore images by adaptively assigning weights to features from different scales according to the blur distribution and content information of the input images. DMTNet combines the advantages of transformer and CNN, in which the vision transformer improves the performance ceiling of CNN, and the inductive bias of CNN enables transformer to extract more robust features without relying on a large amount of data. DMTNet might be the first attempt to use vision transformer to restore the blurring images to clarity. By combining with CNN, the vision transformer may achieve better performance on small datasets. Experimental results on the popular benchmarks demonstrate that our DMTNet significantly outperforms state-of-the-art methods. 
	
\end{abstract}

\noindent 
\section{Introduction} 
In modern cameras, dual-pixel sensors (photodiodes) are used for autofocus~\cite{herrmann2020learning, jang2015sensor, sliwinski2013simple}. When the photoelectric signals of the left and right photodiode is coincident, it indicates that the object is on the depth of field (DoF) of the camera and the image is clear. Conversely,  when the signals from the left and right sensors are phase shifted, defocus blur will appear in the photos.
Defocus blur may affect the performance of subsequent computer vision tasks. For example, in image semantic or instance segmentation tasks, pixels in the defocus blurring region cannot be segmented correctly. 
Especially in the driverless cars, if lane line or freespace cannot be detected correctly in the defocus blurring region, it will lead to the wrong decision in the vehicle identification system and threaten the safety of passengers. Therefore, defocus deblurring is a fundamental and necessary research to avoid the above problems. 

\begin{figure}[t]
	\centering
	\includegraphics[width=1.0\linewidth]{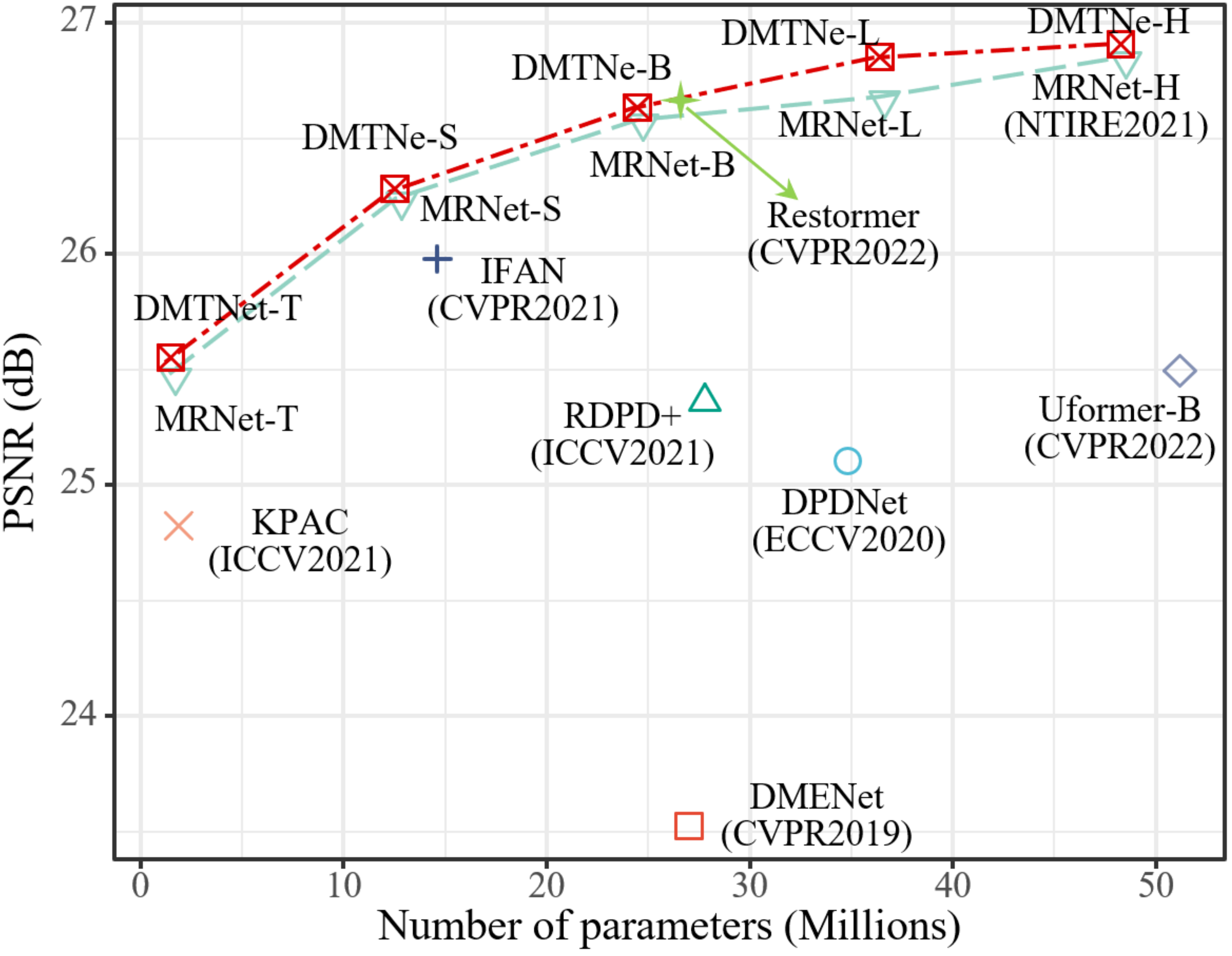}
	\caption{The results of defocus deblurring on the Canon DP dataset~\cite{abuolaim2020defocus}. Under different parameter capacities, our DMTNet achieves the best performance than existing works.} 
	\label{fig:PSNR_SSIM}
\end{figure}

A conventional defocus deblurring method first estimates the defocus blurring map, and then recovers the blurring image to clarity according to the blurring map, which is called two-stage method. The performances of the two-stage methods are affected by the results of defocus blurring map estimation and deblurring algorithm simultaneously. Abuolaim~\etal~\cite{abuolaim2020defocus} propose an end-to-end defocus deblurring convolution neural network (DPDNet) without blurring map estimation. At the same time, they propose a defocus deblurring dataset, named Canon DP data, based on the dual-pixel (DP) camera, which includes left and right defocus blurring images and corresponding sharp image. Compared with the two-stage methods, DPDNet achieves the best performance on Canon DP dataset and takes less time.
However, the blurring image and the sharp image in Cannon DP data are taken under different camera configuration, and there is a problem that the input blurring images are not aligned with the ground truth. To solve this problem, Abuolaim~\etal~\cite{abuolaim2020learning} further explore the causes of defocus blur in the modern cameras. They use standard computer graphics-generated imagery to generate a more realistic synthetic DP dataset, in which the input defocus blurring images and ground truth are aligned. They train a new recurrent convolution network, named RDPDNet, with the synthetic DP dataset to improve the performance. 

DPDNet and RDPDNet are the current state-of-the-art solutions. They all use convolution neural networks (CNN) as the basic operation to restore the blurring image to clear. Thanks to the powerful inductive bias of CNN, both DPDNet and RDPDNet have achieved exciting results on Canon DP dataset. The hard inductive biases of CNN improves the performance of sample-efficient learning, while local constraints of inductive biases lowers the performance ceiling~\cite{d2021convit}. It means that the CNN has bad generalization and robustness on unknown data. Vision transformer breaks the local constraints of CNN, improves feature extraction capabilities, and significantly exceeds the performance of CNN in multiple vision tasks. However, the vision transformer needs to be trained on large scale dataset to give full play to its advantages~\cite{dosovitskiy2020image}. 

Another significant drawback of DPDNet and RDPDNet is that a neural network with the same parameters and architecture is used to restore the blurring images with different blur distribution or content information, which will cause serious problems. The deblurring performance is satisfactory on images with a specific blur distribution, but the robustness to other distribution is very bad. This is why DPDNet and RDPDNet have poor generalization and performance. Collecting a large amount of real data may alleviate the problem, but it is extremely expensive.

In order to obtain better deblurring performance and improve the robustness of the model, this paper proposes a dynamic multi-scale network, named DMTNet, for dual-pixel images defocus deblurring. DMTNet mainly contains two modules: feature extraction module and reconstruction module. The feature extraction module is composed of several vision transformer blocks, which uses its powerful feature extraction capability to obtain richer features and improve the robustness of the model. The reconstruction module is composed of several Dynamic Multi-scale Sub-reconstruction Module (DMSSRM). DMSSRM can restore images by adaptively assigning weights to features from different scales according to the blur distribution and content information of the input images. Compared with the DPDNet or RDPDNet with the fixed weights and architecture, the adaptive dynamic network can significantly improve the performance of deblurring. The feature extraction module based on the vision transformer is used to extract robust features, while the reconstruction module is composed of CNN to restore the blurring images to clarity. Therefore, our DMTNet takes advantage of both the powerful feature extraction capability of vision transformer and the inductive bias capability of CNN. The transformer improves the performance ceiling of CNN, and the inductive bias property of CNN enables transformer to extract more robust features without relying on a large amount of data.

We demonstrate the effectiveness of our DMTNet in dual-pixel defocus deblurring dataset (Canon DP data), which achieves the state-of-the-art performance. As shown in Figure~\ref{fig:PSNR_SSIM}, we significantly improve the performance of defocus deblurring on the Canon DP testing data without using additional training data. When we only use the transformer based feature extraction module to restore the defocus blurring images, our method achieves comparable performance to SOTA method (DPDNet), while number of parameters are reduced by 96.44\%. Especially, when the number of parameters of our method is comparable to the SOTA method (RDPD+), our DMTNet-B achieves 26.63 dB and 0.834 on PSNR and SSIM respectively, 1.24dB higher than RDPD+.
In the reconstruction module, the number of DMSSRM can be flexibly increased or decreased according to the task and performance requirements. When we use four cascaded DMSSRM to restore the defocus blurring image, the deblurring performance will be further improved. 

Our contributions can be summarized as follows:

\begin{itemize}
	\item We propose a feature extraction module based on vision transformer, which improves the feature extraction ability and robustness of the model;
	\item We design an efficient dynamic multi-scale sub-reconstruction module (DMSSRM) based on CNN, which adaptively selects and fuses features from different scales according to blur distribution and content information of the input images to obtain higher performance.
	\item We propose a defocus blurring image restoration network named DMTNet, which is mainly composed of feature extraction module and reconstruction module. DMTNet combines the strength of the vision transformer and CNN, in which vision transformer improves the performance ceiling of CNN and the inductive bias of CNN enables transformer to extract more robust features without relying on a large amount of data.
	\item Our DMTNet achieves state-of-the-art performance on Canon DP data without any additional dataset and significantly outperforms existing works. We also implement several ablation studies to verify the effectiveness of our DMTNet.
\end{itemize}

\section{Related Work} 
\subsection{Defocus deblurring}
Defocus deblurring methods can be divided into two categories: a two-stage method~\cite{d2016non, karaali2017edge, lee2019deep, mao2016image, shi2015just, tang2019defusionnet} using defocus blurring map to guide deblurring and an end-to-end one-stage method~\cite{abuolaim2020defocus, abuolaim2020learning, pan2021dual}. In the two-stage method, in order to generate the defocus blurring map, Yi~\etal~\cite{yi2016lbp} transform the blur detection problem into a segmentation problem. They propose a sharpness metric based on LBP (local binary patterns) and a robust segmentation algorithm to separate in- and out-focus image regions. Shi~\etal~\cite{shi2015just} directly establish correspondence between sparse edge representation and blur strength estimation via the proposed a robust and effective blur feature to generate the defocus blurring map. 
Lee~\etal~\cite{lee2019deep} propose the first end-to-end convolutional neural network (CNN) architecture to estimate the defocus blurring map and achieve the state-of-the-art performance than the traditional methods. 
After getting the defocus blurring map, they use the deconvolution approaches~\cite{levin2011understanding, fish1995blind, krishnan2009fast} to recover shaper information from the defocus blurring image.

The two-stage defocus deblurring methods heavily rely on the accuracy of defocus blurring map, which limits the performance of deblurring. To avoid the influence of defocus blurring map on deblurring performance, Abuolaim~\etal~\cite{ abuolaim2020defocus} propose an end-to-end defocus deblurring framework (DPDNet). DPDNet is U-Net-like neural network architecture. 
In order to train DPDNet, they propose a defocus deblurring dataset, named Canon DP data, based on the dual-pixel camera, which includes left and right defocus blurring images and corresponding sharp image. And DPDNet achieves the best performance on Canon DP dataset and takes less time than the traditional methods. However, the Canon DP data has the problem that the input images and the ground true are not aligned. Abuolaim~\etal~\cite{abuolaim2020learning} further generate a more realistic synthetic DP dataset to solve the misalignment, in which the input defocus blurring images and ground truth are aligned. They train a new recurrent convolution network, named RDPDNet, with the synthetic and Canon DP dataset to improve the performance. The RDPDNet is modified from DPDNet, in which the number of channels are reduced to half at each block to speed up the inference time and convLSTM units are added to learn temporal dependencies present in the image sequence.

\subsection{Vision transformer}
Since the success of AlexNet~\cite{krizhevsky2012imagenet, lecun1989backpropagation} on ImageNet Large Scale Visual Recognition Challenge (ILSVRC)~\cite{deng2009imagenet}, convolutional neural network has been applied to various computer vision tasks.
For example, high level vision tasks include: classification~\cite{simonyan2014very, he2016deep, szegedy2015going}, object detection~\cite{girshick2015fast, ren2015faster, liu2016ssd, redmon2016you} and segmentation~\cite{long2015fully, chen2018encoder}. The low level vision tasks include: image deblurring~\cite{zamir2021multi, tao2018scale}, image super segmentation~\cite{dong2014learning, lim2017enhanced} and image denoising~\cite{brooks2019unprocessing}. And thanks to the powerful inductive bias of CNN, deep learning has achieved exciting results in computer vision tasks. 

VIT~\cite{dosovitskiy2020image} is the first work that uses transformer in computer vision tasks. It surpasses performance of CNN in classification task by virtue of self-attention and large sacle datasets. However, the computational efficiency of self-attention suffers from the high resolution of images. To address above problem, Swin Transformer~\cite{liu2021swin} proposes a shifted windowing scheme in the self-attention, which improves the efficiency by limiting self-attention computation to non-overlapping local windows. It is worth mentioning that the effectiveness of Swin Transformer has been verified in a number of visual tasks, including image classification, semantic segmentation~\cite{lin2021ds}, object detection and video action recognition~\cite{liu2021video}. In addition, some works have also verified the effectiveness of transformer in low vision tasks. Chen~\etal~\cite{chen2021pre} propose a pre-trained transformer model (IPT) with multi-heads and multi-tails to restore the image with the noise, raining and low-resolution. 
Google brain team~\cite{kumar2021colorization} develop a novel approach for diverse high fidelity image colorization based on conditional transformer, named Colorization Transformer (ColTran). 
The above works demonstrate that the transformer can achieve better performance in low-level vision tasks. 

\begin{figure*}[t]
	\centering
	\includegraphics[width=1.0\linewidth]{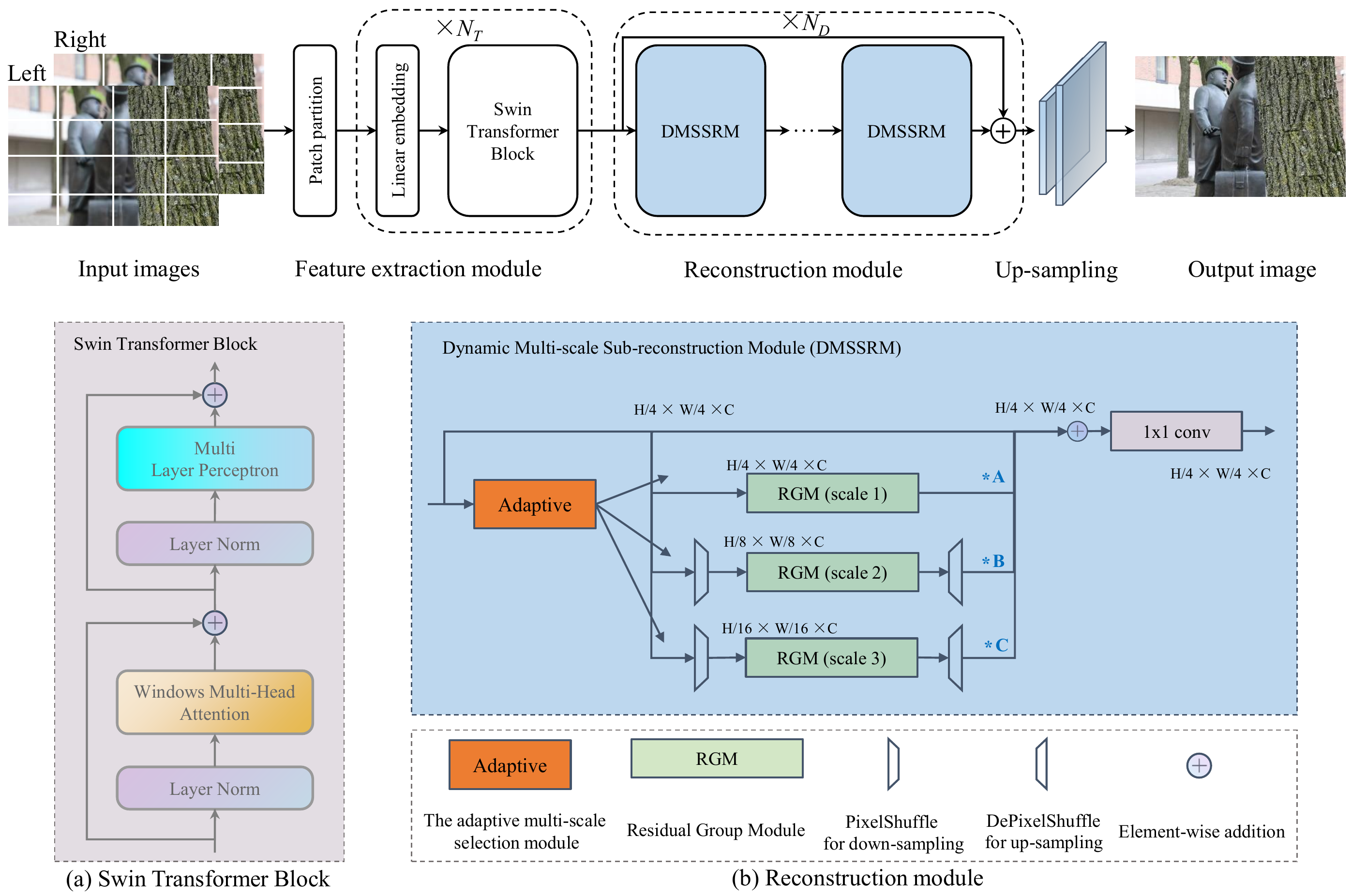}
	\caption{The network architecture of the proposed DMTNet for dual-pixel images defocus deblurring.} 
	\label{fig:Architecture}
\end{figure*}

\section{METHODOLOGY}
\subsection{Network Architecture}
We propose a dynamic multi-scale network, named DMTNet, for dual-pixel images defocus deblurring, as shown in Figure~\ref{fig:Architecture}. DMTNet mainly contains two modules: feature extraction module and reconstruction module. The feature extraction module is composed of several vision transformer blocks, while the reconstruction module is composed of several Dynamic Multi-scale Sub-reconstruction Module (DMSSRM) based on CNN. DMTNet combines the advantages of transformer and CNN, in which the vision transformer improves the performance ceiling of CNN, and the inductive bias of CNN enables transformer to extract more robust features without relying on a large amount of data. In this paper, we input the dual-pixel defocus blurring images ($I_R$ and $I_L$) into DMTNet and output one sharp image ($I_S$).

We concatenate the left $I_L\in\mathbb{R}^{H \times W \times 3}$ and right $I_R\in\mathbb{R}^{H \times W \times 3}$ views of the dual-pixel data to get the input $F_C\in\mathbb{R}^{H \times W \times (2*3)}$ of DMTNet. Then, we send the $F_C$ into the patch partition module to get the token embeddings $F_P$,
\begin{align}
	&F_C = CONCAT(I_R, I_L) \\
	&F_P = D_{Patch}(F_C)
\end{align}
where ($H$, $W$) is the resolution of the input images, 3 and $C$ is the number of channels. $CONCAT$ is the operation that concatenate the right and left views of input dual-pixel on channel dimension.  $D_{Patch}(\cdot)$ denotes patch partition module, which is used to obtain token embeddings as the input of transformer. In the patch partition module, we use 2D convolution operation with the fixed kernel and stride size to obtain the input features $F^{\prime}_P\in\mathbb{R}^{H/P \times W/P \times C}$ with specified patch size $P$. Then we reshape the features $F^{\prime}_P$ into a sequence of flattened 2D patches and layer normalize it to get  $F^{\prime}_{PN}\in\mathbb{R}^{N \times C}$, and $N = WH/P^2$. Then we reverse $F^{\prime}_{PN}$ to get the token embeddings $F_P\in\mathbb{R}^{H/P \times W/P \times C}$. We take the $F_P$ as the input of the feature extraction module constructed by the transformer blocks to extract more robust features $F_E\in\mathbb{R}^{H/P \times W/P \times C}$,
\begin{align}
	F_E = D_{FeatureExtraction}(F_P)
\end{align}
where  $D_{FeatureExtraction}(\cdot)$ represents feature extraction module, which is composed of $N_T$ vision transformer blocks. Inspired by Swin Transformer, self-attention based on the window scheme is used in each transformer block to improve computational efficiency. Then, we send the robust features $F_E$ into the reconstruction module to restore the blurring image to clarity,
\begin{align}
	F_S = D_{Reconstruction}(F_E)
\end{align}
where $F_S\in\mathbb{R}^{H/P \times W/P \times C}$ represents sharp features from reconstruction module, and $D_{Reconstruction}(\cdot)$ denotes reconstruction module, which contains $N_D$ DMSSRM. DMSSRM is a dynamic multi-scale selection network, which dynamically fuses multi-scale features according to the content information of the input images. Finally, sharp features are up-sampled via an up-sampling module to output the sharp image,
\begin{align}
	I_S = D_{Up}(F_S)
\end{align}
where $D_{Up}(\cdot)$ and $I_S\in\mathbb{R}^{H \times W \times 3}$ denote up-sampling module and the final sharp image respectively. We use DePixelShuffle~\cite{vu2018fast, liu2020mmdm} to preserve the information and reduce the parameters in the up-sampling module.

Summarily, $I_S$ can also be represented as follows,
\begin{align}
	I_S = DMTNet(I_R, I_L)
\end{align}
where $DMTNet(\cdot)$ denotes the function of DMTNet.

\subsection{Feature Extraction Module}
The feature extraction module is composed of several vision transformer blocks, which uses its powerful feature extraction capability to obtain robust features. In the standard global self-attention, the computational complexity is quadratic with the number of tokens, making it unsuitable for low-level vision tasks with high-resolution. Inspired by the Swin Transformer, we replaced the standard global attention with the non-overlapping windows based self-attention for performing efficient modeling in low-level vision tasks. The computational complexity formulas for standard multi-head self-attention (MSA) and window-based multi-head self-attention (WMSA) are as follows,
\begin{align}
	& \Omega(MSA) = 4hwC^2 + 2(hw)^2C \\
	& \Omega(WMSA) = 4hwC^2 + 2W^2hwC \\
	& \frac{\Omega(MSA)}{\Omega(WMSA)} \approx \frac{hw}{3W^2} \label{eq:COM}
\end{align}
where ($h$, $w$) is the resolution of the input features, $C$ is the number of channels and $W$ is the size of windows. In this paper, $C \ll hw$ and $C \approx W^2$. According to Eq.\eqref{eq:COM}, the window-based multi-head self-attention significantly reduces the computational complexity of the model. 

As shown in Figure~\ref{fig:Architecture}(a), each transformer block is composed of a window-based multi-head self-attention (WMSA) and a multi-layer perceptron (MLP). We perform layer normalization on the input features of WMSA and MLP modules, add residual connections to alleviate the gradient disappearance. The transformer block is formulated as,
\begin{align}
	& X = WMSA(LN(X)) + X \\
	& X = MLP(LN(X)) + X
\end{align}
where $WMSA(\cdot)$ is the WMSA module, $LN(\cdot)$ is the LayerNorm (LN) layer and $MLP(\cdot)$ is the MLP module. Specifically, $X\in\mathbb{R}^{W^2 \times C}$ is the local window feature.

\subsection{Reconstruction Module}
The reconstruction module is composed of several DMSSRM, as shown in Figure~\ref{fig:Architecture}. Compared with the DPDNet or RDPDNet with the fixed weights and architecture, DMSSRM restores images by adaptively assigning weights to features from different scales according to the blur distribution and content information of the input images. 
The reconstruction module is a multi-stage learning process, which gradually restores the blurring image through the cascaded DMSSRM module. 
\begin{align}
	\begin{split}
		F_{i} = D_{DMSSRM_i}(F_{i-1}), \quad i = 1, 2, \cdots, D,
	\end{split}	
\end{align}
where $ D_{DMSSRM_i}(\cdot)$ indicates the $i$-th DMSRGM module in the reconstruction module. $F_{i}\in\mathbb{R}^{H/P \times W/P \times C}$ represents the features extracted by the $i$-th DMSRGM. $F_0 = F_E$ represents the features extracted by the feature extraction module. Then, we add a global residual connections to speed up the gradient feedback in reconstruction module. The output of reconstruction module is formulated as,
\begin{align}
	\begin{split}
		F_S = F_D + F_0
	\end{split}	
\end{align}
where $F_D$ denotes the reconstructed features of the last DMSRGM.

\subsubsection{Dynamic Multi-scale Sub-reconstruction Module (DMSSRM)}
Both DPDNet and RDPDet adopts U-Net-like neural network architecture to remove the blur in the input images, and improve the deblurring performance by extracting and fusing multi-scale features. However, 
if the neural network with the fixed weights and architecture is used to restore the blurring image with different blurring distribution and content information will lead to the bad robustness in real scene. Therefore, we propose an Dynamic Multi-scale Sub-reconstruction Module (DMSSRM), which can adaptively assign weights to features from different scales according to the blurring distribution and content information of the input images. 

DMSSRM is composed of multi-scale feature extraction module, adaptive multi-scale selection module and fusion module, as shown in Figure~\ref{fig:Architecture}(b). In each scale branch, we use the same configured residual group module (RGM) without Channel Attention to restore the detailed information in the blurring image ~\cite{zhang2018image, liu2020mmdm}. Then, we use the adaptive multi-scale selection module to adaptively fuse multi-scale features according to the blurring distribution and content information of the input images. Finally, we use the fusion module to fuse and exchange information from different scales.
\begin{align}
	\begin{split}
		F_{i} &= D_{DMSSRM_i}(F_{i-1}) \\
		&= W(F_{scale\_1} * \alpha 1 + \cdots + F_{scale\_n}  * \alpha n + F_{i-1})
	\end{split}	
\end{align}
where $D_{DMSSRM_i}(\cdot)$ denotes DMSSRM module. $F_i$ and $F_{i-1}$ are the output and input of $i$-th DMSSRM. $F_{scale\_1}, \cdots, F_{scale\_n}$ represent the features from different scales respectively in DMSSRM, while $\alpha 1, \cdots,\alpha n$ represent the weights for each scale predicted by adaptive multi-scale selection module.  $W$ is the weight of convolution layers with kernel size of 1$\times$1 of fusion module and used to fuse the multi-scale features.

\subsubsection{Adaptive multi-scale selection module}
\begin{figure}[t]
	\centering
	\includegraphics[width=1.0\linewidth]{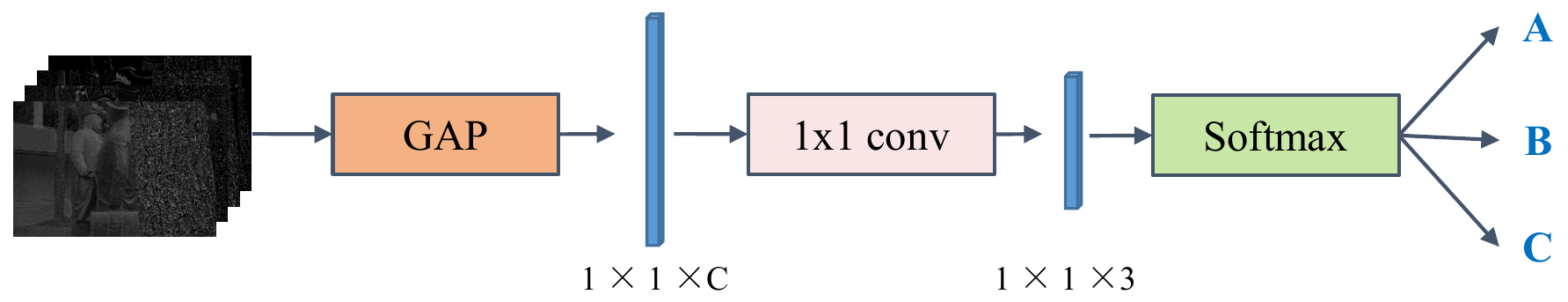}
	\caption{\small
		Adaptive multi-scale selection module.
	}
	\label{fig:ADSSM}
\end{figure}

Adaptive multi-scale selection module is proposed to predict weights for features from each scale based on image contents information and blur distribution. As shown in Figure~\ref{fig:ADSSM}, adaptive multi-scale selection module  is composed of global average pooling layer and convolution operation with kernel size of 1$\times$1, and assigns corresponding weights to the features from each scale through softmax activation function. Global average pooling layer aggregates the input feature maps into a 1$\times$1$\times$C vector, it means that we use features with 1$\times$1$\times$C to represent the blurring distribution and content information of the input features. Then, the weights of each scale are predicted using 1$\times$1 convolution and softmax,
\begin{align}
	F_C = W(GAP(F_{i-1})) \\
	\alpha_j = \frac{\exp(F_{C_j})}{\sum_{j=1}^{S}\exp(F_{C_j})}
\end{align}
where $GAP(\cdot)$ denotes global average pooling,  $W$ is the weight of convolution layers, $\alpha_j$ represents the weight coefficient of the $j$-th scale, $S$ is the number of scales.

\begin{table*}[t]
	\centering
	\scalebox{0.90}{	
		\begin{tabular}{c ccccccccccc}
			\hline			
			\multirow{2}{*}{\bf Method} &\multicolumn{3}{c}{\bf Indoor} &\multicolumn{3}{c}{\bf Outdoor} &\multicolumn{3}{c}{\bf Combined} & \multirow{2}{*}{{\bf Param}$\downarrow$} & \multirow{2}{*}{{\bf Flops}$\downarrow$} \\ \cline{2-10}
			
			& PSNR $\uparrow$ & SSIM $\uparrow$ & MAE $\downarrow$ & PSNR $\uparrow$ & SSIM $\uparrow$ & MAE $\downarrow$ & PSNR $\uparrow$ & SSIM $\uparrow$ & MAE $\downarrow$ \\ \hline
			
			EBDB~\cite{karaali2017edge}	& 25.77 & 0.772 & 0.040 & 21.25 & 0.599 & 0.058 & 23.45 & 0.683 & 0.049 & - & - \\ 		
			
			DMENet~\cite{lee2019deep}	& 25.70 & 0.789 & 0.036 & 21.51 & 0.655 & 0.061 & 23.55 & 0.720 & 0.049 & 26.71M & - \\ 
			
			JNB~\cite{shi2015just}		& 26.73 & 0.828 & 0.031 & 21.10 & 0.608 & 0.064 & 23.84 & 0.715 & 0.048 & - & - \\ 
			
			DPDNet~\cite{abuolaim2020defocus}	& 27.48 & 0.849 & 0.029 & 22.90 & 0.726 & 0.052 & 25.13 & 0.786 & 0.041 & 34.52M & 1883.74G\\ 
			
			DPDNet+~\cite{abuolaim2020defocus}	& 27.65 & 0.852 & 0.028 & 22.72 & 0.719 & 0.054 & 25.12 & 0.784 & 0.042 & 34.52M & 1883.74G \\ 
			
			RDPD+~\cite{abuolaim2021learning}	& 28.10 & 0.843 & 0.027 & 22.82 & 0.704 & 0.053 & 25.39 & 0.772 & 0.040 & 27.51M & 612.05G \\
			
			IFAN~\cite{lee2021iterative}	& 28.66 & 0.868 & 0.025 & 23.46 & 0.743 & 0.049 & 25.99 & 0.804 & 0.037 & 14.48M & 855.31G \\
			
			KPAC~\cite{son2021single}	& 27.36 & 0.849 & 0.030 & 22.47 & 0.707 & 0.055 & 24.85 & 0.776 & 0.043 & {1.58M} & {348.63G} \\
			
			Uformer-B~\cite{wang2021uformer}	& 28.23 & 0.860 & 0.026 & 23.10 & 0.728 & 0.051 & 25.65 & 0.795 & 0.039 & 50.88M & 3066.94G \\ 
			
			Restormer~\cite{zamir2022restormer}	& 29.48 & 0.895 & 0.023 & 23.97 & 0.773 & 0.047 & 26.66 & 0.833 & 0.035 & 26.13M & 4059.08G \\ 
			
			MRNet~\cite{abuolaim2021ntire}	& 29.53 & 0.896 & 0.022 & 24.31 & 0.789 & 0.045 & 26.85 & 0.841 & 0.034 & 48.25M  & 2597.38G\\ \hline \hline
			
			DMTNet-T	& 27.73 & 0.850 & 0.027 & 23.38 & 0.724 & 0.051 & 25.50 & 0.786 & 0.039 & 1.23M & 147.37G \\
			
			DMTNet-S	& 28.77 & 0.880 & 0.025 & 23.90 & 0.765 & 0.046 & 26.27 & 0.821 & 0.036 & 12.57M & 688.35G \\
			
			DMTNet-B	& 29.21 & 0.891 & {0.023} & 24.18 & 0.780 & {0.046} & 26.63 & 0.834 & {0.035} & 24.46M & 1294.41G \\
			
			DMTNet-L & {29.50} & 0.892 & {0.023} & {24.31} & {0.783} & {0.045} & {26.83} & {0.836} & {0.034} & 36.35M & 1900.46G \\ 
			
			DMTNet-H	& {29.50} & {0.895} & {0.022} & {24.45} & {0.791} & {0.045} & {26.91} & {0.841} & {0.034} & 48.24M  & 2506.51G\\ \hline
			
		\end{tabular}
	}
	\caption{Dual-pixel defocus deblurring comparison on Canon DP dataset with PSNR, SSIM and MAE. DPDNet (ECCV2020), IFAN (CVPR2021), KPAC (ICCV2021), Uformer-B (CVPR2022), Restormer (CVPR2022), DMTNet-T, DMTNet-S, DMTNet-B, DMTNet-L and DMTNet-H are trained on Canon DP data. DMTNet-S, DMTNet-B, DMTNet-L and DMTNet-H use 1, 2, 3, 4 DMSSRM as the reconstruction module respectively, while DMTNet-T without using DMSSRM. DPDNet+ and RDPD+ (ICCV2021) are trained with Canon and synthetic DP data. Compared to the other SOTA methods, our DMSSRM more effectively and efficiently removes defocus blurring by using the dual-pixel images.}	
	\label{tab:quantitaiveResultsCanon}
\end{table*}

\section{EXPERIMENTS}

\subsection{Dataset}
We train and test DMTNet on Canon DP datase without using additional data. Abuolaim~\etal~\cite{abuolaim2020defocus} collecte 500 pairs of blurring and clear images using the dual-pixel camera, in which each pair contains two blurring images and one clear image. Following DPDNet, we divide the data into 70\% training, 15\% validation, and 15\% testing sets. The resolution of each image is 1680 $\times$1120. To speed up the training, we slide a window with a size of 512$\times$512 and 60\% overlap in the training sets to extract more image patches. Data augmentation is performed on training data, which includes horizontal and vertical flips. 

\subsection{Implementation Details}
DMTNet is composed of a patch partition module, a feature extraction module, a reconstruction module and an up-sampling module. In the patch partition module, we set the patch size to 4$\times$4. The feature extraction module is composed of 5 transformer blocks, and the number of channels is the same as patch partition module, which is 96. The reconstruction module consists of 4 DMSSRM. In DMSSRM, we use residual group module (RGM) to extract features on each scale. The residual module (RM) is composed of two 3$\times$3 convolution and $PReLU$ activation functions, the number of channels is 64. And we use 10 RM to construct residual block module (RBM) and use 5 RBM to form residual group module (RGM). The initial learning rate is 1e-4, and we use the cosine annealing learning rate scheduler~\cite{loshchilov2016sgdr} with about 1,000 epochs. We use the Adam~\cite{kingma2014adam} with $\beta_1=0.9$ and $\beta_2=0.999$ to optimize the Charbonnier loss function~\cite{lai2018fast}. We use PyTorch 1.6, NVIDIA V100 GPU with CUDA11.0 to accelerate training.

\begin{figure*}[!t]
	\begin{center}
		\includegraphics[width=1.0\linewidth]{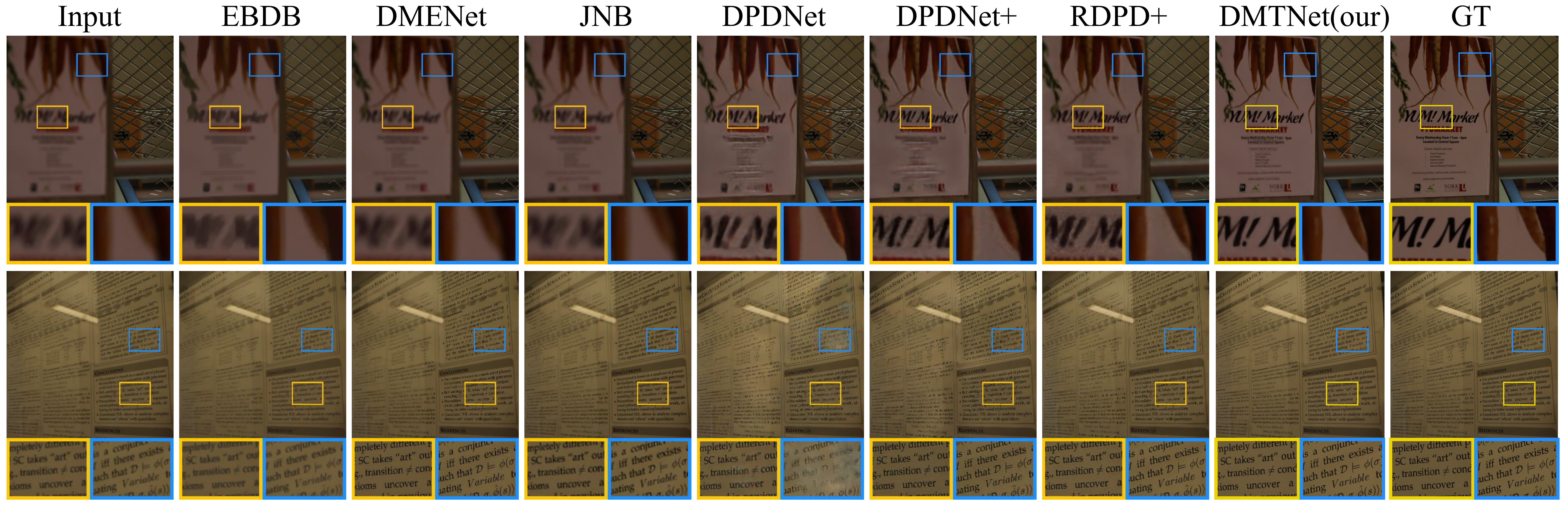}
		\caption{Qualitative comparisons with different motion deblurring methods on Canon DP testing dataset. The first column is the input blurring image, and the last column is the corresponding ground truth sharp image. The columns in between are the results of different methods. We present zoomed-in cropped patches in yellow and blue boxes. Compared to the state-of-the-art methods, the images restored by DMTNet are sharper.}
		\label{fig:canno_dp}
	\end{center}
\end{figure*}

\subsection{Evaluation and Comparison}
\subsubsection{Comparisons with State-of-the-art Methods}
The quantitative results of our DMTNet on dual-pixel defocus deblurring task achieve SOTA performance compared to other models, which is present in Table~\ref{tab:quantitaiveResultsCanon}. EBDB~\cite{karaali2017edge}, DMENet~\cite{lee2019deep} and JNB~\cite{shi2015just} are two-stage defocus deblurring algorithms, which first estimate the defocus map and then use the generated defocus map to guide the deblurring process. EBDB and JNB are traditional defocus blurring detection algorithms, while DMENet is a method based on deep learning. They are combined with non-blind deblurring algorithms~\cite{krishnan2009fast, fish1995blind} to remove the blurring in the images. DPDNet~\cite{abuolaim2020defocus}, DPDNet+ and PDPD+~\cite{abuolaim2020learning} are end-to-end algorithms, and they are all trained on dual-pixel datasets. DPDNet is trained on Canon DP training data, while DPDNet+ and PDPD+ are trained on both Canon and synthetic DP data. Our DMTNet-B is trained only on the Canon DP Dataset without using additional synthetic data. DMTNet-B indicates that using 2 DMSSRM as the reconstruction module, while DMTNet-H uses 4 DMSSRM.  We divide the test data into three scene categories: indoor, outdoor and combined. Our DMTNet significantly improves defocus deblurring performance and achieves best results for all metrics. Specifically, when the number of parameters of our DMTNet is comparable to the SOTA method, our DMTNet-B achieves 29.21, 24.18 and 26.63 dB on three scenarios respectively, being 4.0\%, 6.4\% and 4.9\% better than the previous method (RDPD+). When using more DMSSRM to restore defocus blurring images, our performance is further improved. MRNet~\cite{abuolaim2021ntire} won the first place on the NTIRE 2021 Challenge for Defocus Deblurring Using Dual-pixel Images (@CVPR 2021). However, our DMTNet-H achieves better performance and fewer Flops than MRNet. 
The visual results are presented in Figure~\ref{fig:canno_dp}, while the images restored by our DMTNet are clearer.

\subsection{Ablation Study}
\subsubsection{Impact of DMSSRM number}
In the reconstruction module, we explore the influence of the number of DMSSRM on deblurring performance. In fact, the later DMSSRM module is used to relearn and refine the previously fused features. As shown in Figure ~\ref{fig:PSNR_SSIM} and Table ~\ref{tab:quantitaiveResultsCanon}, the PSNR is positively correlated with numbers of DMSSRM, while the total number of parameters increases gradually. When we only use the transformer based feature extraction module (reconstruction module is not used) to restore the defocus blurring images, our method achieves comparable performance to SOTA method (DPDNet), while number of parameters are reduced by 96.44\%. When we use one DMSSRM as the reconstruction module, we significantly improve the deblurring performance. And combining with the feature extraction module, we improve the PSNR by 0.77dB, proving the effectiveness of our DMSSRM. Especially, when the number of parameters of our method is comparable to the SOTA method, our DMTNet achieves 26.91 dB on PSNR, being 0.06dB higher than SOTA method and significantly improving the deblurring performance.


\subsubsection{Impact of feature extraction module with transformer}
Our DMTNet is a Transformer-CNN strategy, which combines advantage of both the powerful feature extraction capability of vision transformer and the inductive bias capability of CNN. We also perform ablation experiments on Canon DP dataset to verify the effectiveness of the strategy. The experimental results are shown in Table~\ref{tab:Transformer_Dynamic}. In the non-transformer module (or named CNN based module), we use two convolutional neural networks with kernel size of 3$\times$3 and stride size of 2 replace the transformer block for extracting the features. As shown in Table~\ref{tab:Transformer_Dynamic}, the transformer based feature extraction module (method 3) is 0.21dB better than the CNN based module (method 1). The experimental results also proves that the transformer can extract robust features without relying on a large amount of dataset combining with CNN.

\subsubsection{Impact of the dynamic multi-scale on DMSSRM}
\begin{figure}[!t]
	\begin{center}
		\includegraphics[width=1.0\linewidth]{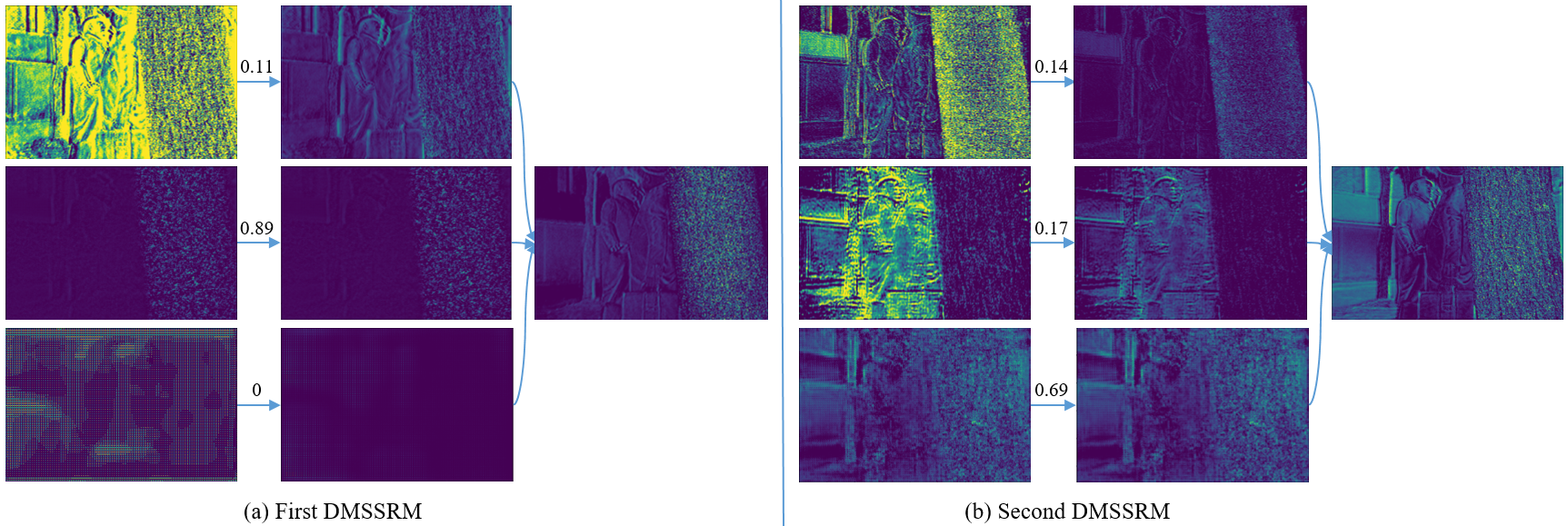}
		\caption{Feature visualization of DMSSRM on different scales.}
		\label{fig:0009}
	\end{center}
\end{figure}

We visualize the features and weights of the two DMSSRM, as shown in Figure~\ref{fig:0009}. In the first DMSSRM, scale 1 enhances the texture information of features, but overexposure occurs. Therefore, it is adjusted by the adaptive module to make the texture clearer. Scale 2 is responsible for preserving the features of clear areas. Scale 3 contains a lot of noise information, so it is assigned the minimum weight (1.1043e-07$\approx$0.0). In the second DMSSRM, Scale 1 is mainly responsible for enhancing and maintaining the texture of clear areas. Scale 2 is responsible for restoring the features of smooth and blurred areas. Scale 3 restores the entire image from a global perspective. Figure~\ref{fig:0009} qualitatively demonstrate that DMSSRM can restore images by adaptively assigning weights to features from different scales according to the content information of the feature maps. 

Dynamic multi-scale selection is leitmotif of our DMSSRM. We also conduct an ablation study about dynamic multi-scale to quantitatively demonstrate that dynamic multi-scale network can achieve better performance than fixed weights and architecture. Compared with the fixed weights and architecture, DMSSRM restore images by adaptively assigning weights to features from different scales according to the blur distribution and content information of the input images. We remove the adaptive multi-scale selection module in DMSSRM as the fixed architecture. As shown in Table~\ref{tab:Transformer_Dynamic}, DMSSRM with dynamic multi-scale selection (method 3) achieves 26.63dB on Canon DP dataset, and is 0.16dB better than the fixed architecture (method 2).

\begin{table}[t]
	\centering	
	\scalebox{1.0}{	
		\begin{tabular}{cccccc}
			\hline						
			Method	& Transformer & Dynamic & PSNR  & SSIM  \\ \hline
			
			1 &               &  $\checkmark$ & 26.42 & 0.830 \\  
			
			2 & $\checkmark$  &               & 26.47 & 0.832 \\  
			
			3 & $\checkmark$  &  $\checkmark$ & {\bf 26.63} & {\bf 0.834}\\ \hline
		\end{tabular}
	}	
	\caption{Comparison on Canon DP dataset with/without transformer module and adaptive multi-scale selection module (using $\times$2 DMSSRM for quick verification).}
	\label{tab:Transformer_Dynamic}
\end{table}

\section{CONCLUSION}
In this paper, we propose a Transformer-CNN combined defocus deblurring model DMTNet. And DMTNet might be the first attempt to use vision transformer to restore the blurring images to clarity. DMTNet mainly contains two modules: feature extraction module and reconstruction module. The feature extraction module is composed of several vision transformer blocks, which is used to extract the robust features. The reconstruction module is composed of several Dynamic Multi-scale Sub-reconstruction Module (DMSSRM). DMSSRM can restore images by adaptively assigning weights to features from different scales according to the blur distribution and content information of the input images. Compared with the fixed architecture, the adaptive dynamic network can significantly improve the performance of deblurring. Our DMTNet combines advantage of both the powerful feature extraction capability of vision transformer and the inductive bias capability of CNN. The transformer improves the performance ceiling of CNN, and the inductive bias of CNN enables transformer to extract more robust features without relying on a large amount of data.  Experimental results on popular benchmark demonstrate that our DMTNet significantly outperforms existing solutions and achieves state-of-the-art performance withou using additional synthetic data. 

{\small
\bibliographystyle{ieee_fullname}
\bibliography{main}
}

\end{document}